\documentclass[letterpaper]{article} % DO NOT CHANGE THIS
\usepackage[submission]{aaai25}  % DO NOT CHANGE THIS
\usepackage{times}  % DO NOT CHANGE THIS
\usepackage{helvet}  % DO NOT CHANGE THIS
\usepackage{courier}  % DO NOT CHANGE THIS
\usepackage[hyphens]{url}  % DO NOT CHANGE THIS
\usepackage{graphicx} % DO NOT CHANGE THIS
\usepackage{amsmath} %添加这个包以支持高级数学公式
\usepackage{comment}
\usepackage{natbib}  % DO NOT CHANGE THIS AND DO NOT ADD ANY OPTIONS TO IT
\usepackage{caption} % DO NOT CHANGE THIS AND DO NOT ADD ANY OPTIONS TO IT
\usepackage{algorithm}
\usepackage{algorithmic}
\usepackage{booktabs}
\usepackage{multirow}
\usepackage{newfloat}
\usepackage{listings}
\DeclareCaptionStyle{ruled}{labelfont=normalfont,labelsep=colon,strut=off} % DO NOT CHANGE THIS
\floatstyle{ruled}
\newfloat{listing}{tb}{lst}{}
\floatname{listing}{Listing}
\title{COEFF-KANs: A Paradigm to Address the Electrolyte Field with KANs}
\author{
    Xinhe Li\textsuperscript{\rm 1}\textsuperscript{\rm 3}\equalcontrib, Zhuoying Feng\textsuperscript{\rm 1}\textsuperscript{\rm 3}\equalcontrib, Yezeng Chen\textsuperscript{\rm 2}\textsuperscript{\rm 3}\equalcontrib, Weichen Dai\textsuperscript{\rm 1}\textsuperscript{\rm 3}, Zixu He\textsuperscript{\rm 1}, \\ 
    Yi Zhou\textsuperscript{\rm 1}\textsuperscript{\rm 3}\corr, Shuhong Jiao\textsuperscript{\rm 1}\corr
}
\affiliations{
    \textsuperscript{\rm 1}University of Science and Technology of China\\
    \textsuperscript{\rm 2}Shanghaitech University\\
    \textsuperscript{\rm 3}Knowledge Computing Lab\\
    lixinhe669@gmail.com, zoeyfeng@mail.ustc.edu.cn, chenyz2022@shanghaitech.edu.cn, dddwc@mail.ustc.edu.cn, hzx123@mail.ustc.edu.cn, yi\_zhou@ustc.edu.cn, jiaosh@ustc.edu.cn
}

\newcommand*{\OURMODEL}{COEFF}
\newcommand*{\OURMODELKAN}{COEFF-KAN}
\newcommand*{\OURMODELMLP}{COEFF-MLP}
\newcommand*{\TESTIN}{$\mathcal{T}_\text{in}$}
\newcommand*{\TESTOUT}{$\mathcal{T}_\text{out}$}

\usepackage{bibentry}
\begin{document}

\maketitle

\begin{abstract}

To reduce the experimental validation workload for chemical researchers and accelerate the design and optimization of high-energy-density lithium metal batteries, we aim to leverage models to automatically predict Coulombic Efficiency (CE) based on the composition of liquid electrolytes. There are mainly two representative paradigms in existing methods: machine learning and deep learning. However, the former requires intelligent input feature selection and reliable computational methods, leading to error propagation from feature estimation to model prediction, while the latter (e.g. MultiModal-MoLFormer) faces challenges of poor predictive performance and overfitting due to limited diversity in augmented data. To tackle these issues, we propose a novel method \OURMODEL{} (COlumbic EFficiency prediction via Fine-tuned models), which consists of two stages: pre-training a chemical general model and fine-tuning on downstream domain data. Firstly, we adopt the publicly available MoLFormer model to obtain feature vectors for each solvent and salt in the electrolyte. Then, we perform a weighted average of embeddings for each token across all molecules, with weights determined by the respective electrolyte component ratios. Finally, we input the obtained electrolyte features into a Multi-layer Perceptron or Kolmogorov-Arnold Network to predict CE. Experimental results on a real-world dataset demonstrate that our method achieves SOTA for predicting CE compared to all baselines. Data and code used in this work will be made publicly available after the paper is published.

\end{abstract}

% Uncomment the following to link to your code, datasets, an extended version or similar.
%
% \begin{links}
%     \link{Code}{https://aaai.org/example/code}
%     \link{Datasets}{https://aaai.org/example/datasets}
%     \link{Extended version}{https://aaai.org/example/extended-version}
% \end{links}

\section{Introduction}

% 研究任务
% 1. 电解液配比设计的作用
% 2. 高通量筛选方法的局限性
% 3. 引出任务（介绍输入和输出），电解液配比和库伦效率
The discovery and optimization of battery electrolytes are crucial in climate and sustainable technology due to their potential to accelerate decarbonization in all economic sectors~\cite{climate1,climate2}. 
Enhancing Coulombic Efficiency (CE) is vital for the adoption of high-energy-density lithium metal batteries~\cite{ce1,ce2}. 
Liquid electrolyte engineering is a promising strategy for improving CE, but its complexity makes it challenging to predict and design electrolyte performance~\cite{engineering}. 
For example, high-throughput screening accelerates the search for individual compounds but does not guide the comprehensive design of new formulations~\cite{high-throughput-screening}. 
Fortunately, research has shown that modern energy storage devices typically use liquid electrolytes composed of one or more organic solvents and salt additives, and their formulations have a significant impact on CE~\cite{task1,task2}.
Thus, we can predict CE among a large number of electrolyte formulations and screen out those with high CE to recommend to chemical researchers, significantly reducing the workload of experimental validation.
% Thus, our task is to predict CE based on electrolyte components and recommend a set of formulations to chemical researchers, significantly reducing the workload of experimental validation.

\begin{figure}[t]
    \centering
    \includegraphics[width=\linewidth]{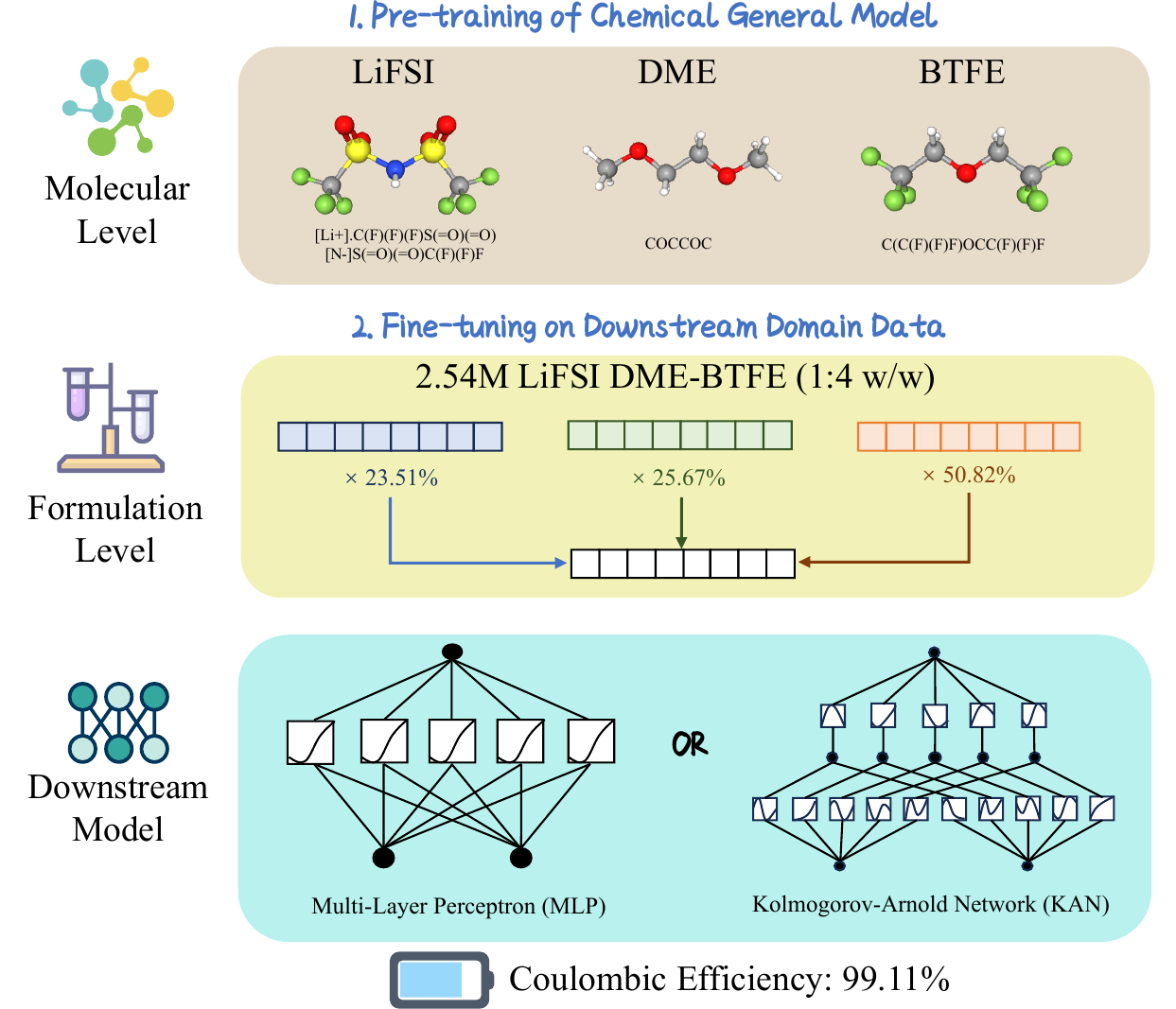}
    \caption{Overview of our method's two-stage paradigm: pre-training a chemical general model and fine-tuning on downstream domain data.} 
    \label{fig:intro}
\end{figure}

% 已有方法 + 挑战
To deal with this challenge, existing works are mainly based on two representative paradigms: machine learning~\cite{kim,data-driven} and deep learning methods~\cite{uni-elf,multimodal-moLformer, molformer, f-gcn}.
The former approaches predict the CE of electrolytes via a supervised learning model such as regression, which takes selective features of components (e.g., redox potentials, dielectric constants, ionic conductivity, and elemental composition) as input.
This process requires intelligent input feature selection and reliable computational methods, which raises the critical issue of error propagation from estimating input features to model predictions (e.g., \citet{kim} ignore the structural information of chemical molecules).
% KAN 如果能做拟合，可以把目前模型的可解释性差加上
An exemplary work in the latter approaches is MultiModal-MoLFormer~\cite{multimodal-moLformer}, which leverages extensive chemical information learned from unlabeled corpora in pre-training to predict the CE of electrolytes. 
However, its predictive performance remains suboptimal, with a Root Mean Square Error (RMSE) of only 0.195 on the test set. 
Furthermore, data augmentation by shuffling the order of electrolyte components effectively increased the dataset size from 147 to 27,266 samples, yet limited diversity may lead to overfitting during training.

% The former approach uses the elemental composition of liquid electrolytes as features for machine learning models (e.g., linear regression, support vector machines, bagging, and random forests). 
% They successfully reveal that a reduction in the solvent oxygen content is critical for superior CE but overlook structural information of chemical molecules (e.g., geometric arrangements of elements in isomers). 
% This subtle distinction has been shown to affect electrolyte properties and battery performance.

% 我们的方法
To address the above issues, we propose a novel electrolyte property prediction method \OURMODEL{} (COlumbic EFficiency prediction via Fine-tuned models).
As illustrated in Figure~\ref{fig:intro}, \OURMODEL{} is intensively designed for learning the relationship between electrolyte composition and CE, which consists of two stages: pre-training of chemical general models and fine-tuning on downstream domain data.
First, we adopt the publicly available MoLFormer~\cite{molformer}, which is constructed by pre-training an efficient Transformer~\cite{transformer} encoder on a large unlabeled chemical corpus to obtain embeddings at the molecular level.
% The dataset is sourced from a subset combining 10\% of the Zinc and 10\% of the PubChem databases.
The input of MoLFormer conforms to the Simplified Molecular Input Linear Entry Specification (SMILES), which can represent both the elemental composition and structural information of chemical molecules.
For our application, we independently input all solvents and salts in the electrolyte into MoLFormer to extract the chemical features of each component.
Then, we multiply the token embeddings of all SMILES molecules by their corresponding ratios and sum them to obtain representations at the formulation level.
This is similar to concatenating all embeddings along the token length dimension, followed by a weighted average pooling layer, with the weights derived from the ratios of the electrolyte components.
Finally, we feed electrolyte embeddings into a downstream model, including Multi-layer Perceptron (\OURMODELMLP{}) or Kolmogorov–Arnold Networks (\OURMODELKAN{})~\cite{kan}, and the output is the predicted value of CE.
Our work demonstrates the potential of deep learning in electrolyte design. By harnessing the power of language models, we can accelerate the discovery and optimization of new formulations, significantly reducing the experimental validation workload for chemical researchers, and potentially revolutionizing various industries.

% 我们构建的数据集

% 实验结果
Experimental results on a real-world test set with 13 data points show that our method (Min. 0.132 for \OURMODELMLP{} and Min. 0.110 for \OURMODELKAN{}) achieves significant and consistent improvement in the RMSE metric for predicting CE compared to all baselines, including machine learning (Min. 0.170) and deep learning (Min. 0.195) methods.
% 其他的实验结果
We will publish all source codes and datasets of this work on GitHub for further research explorations.

% 贡献点
In summary, the main contributions of this paper are threefold:
(1) We focus on the task of predicting CE based on electrolyte components and propose a novel method \OURMODEL{}, which consists of two stages: pre-training a chemical general model and fine-tuning on downstream domain data.
% leverages the pre-trained language model MoLFormer and supervised fine-tuning on the downstream task, fully capturing the structural information and elemental composition of electrolytes.
(2) To the best of our knowledge, we are the first to introduce KAN networks into the task of electrolyte property prediction and achieve better performance than MLP. 
% 加新的数据集
(3) We conduct extensive experiments on real-world datasets and show that \OURMODEL{} outperforms previous state-of-the-art works.

\section{Related Work}

\subsection{Language Models for Chemistry}
In recent years, the realm of large language models has seen significant advancements, highlighting their remarkable ability to understand intricate chemical language representations~\cite{llm1,llm2,llm3}. 
Studies have shown that the two-stage paradigm has great potential: first pre-training on large unlabeled datasets and then fine-tuning on a specific downstream task~\cite{two-stages1,molformer,two-stages3}.
Leveraging self-supervised learning, these methods successfully minimize reliance on labeled data and specific tasks. 
While pre-trained language models perform well in predicting molecular properties~\cite{property-1,property-2}, their practical implementation on further downstream tasks is still in its early stages and largely limited to sequencing problems such as predicting protein sequences~\cite{protein-sequences}, polymers~\cite{polymers}, and chemical reactions~\cite{chemical-reactions}.
Moreover, prediction errors of most methods~\cite{multimodal-moLformer,f-gcn,kim} in the electrolyte field are larger than experimental errors, making them difficult to achieve the goal of practical application.
We focus on the rarely studied task of CE prediction and propose a novel method \OURMODEL{} following the two-stage paradigm.

\begin{figure*}[t]
    \centering
    \includegraphics[width=.9\linewidth]{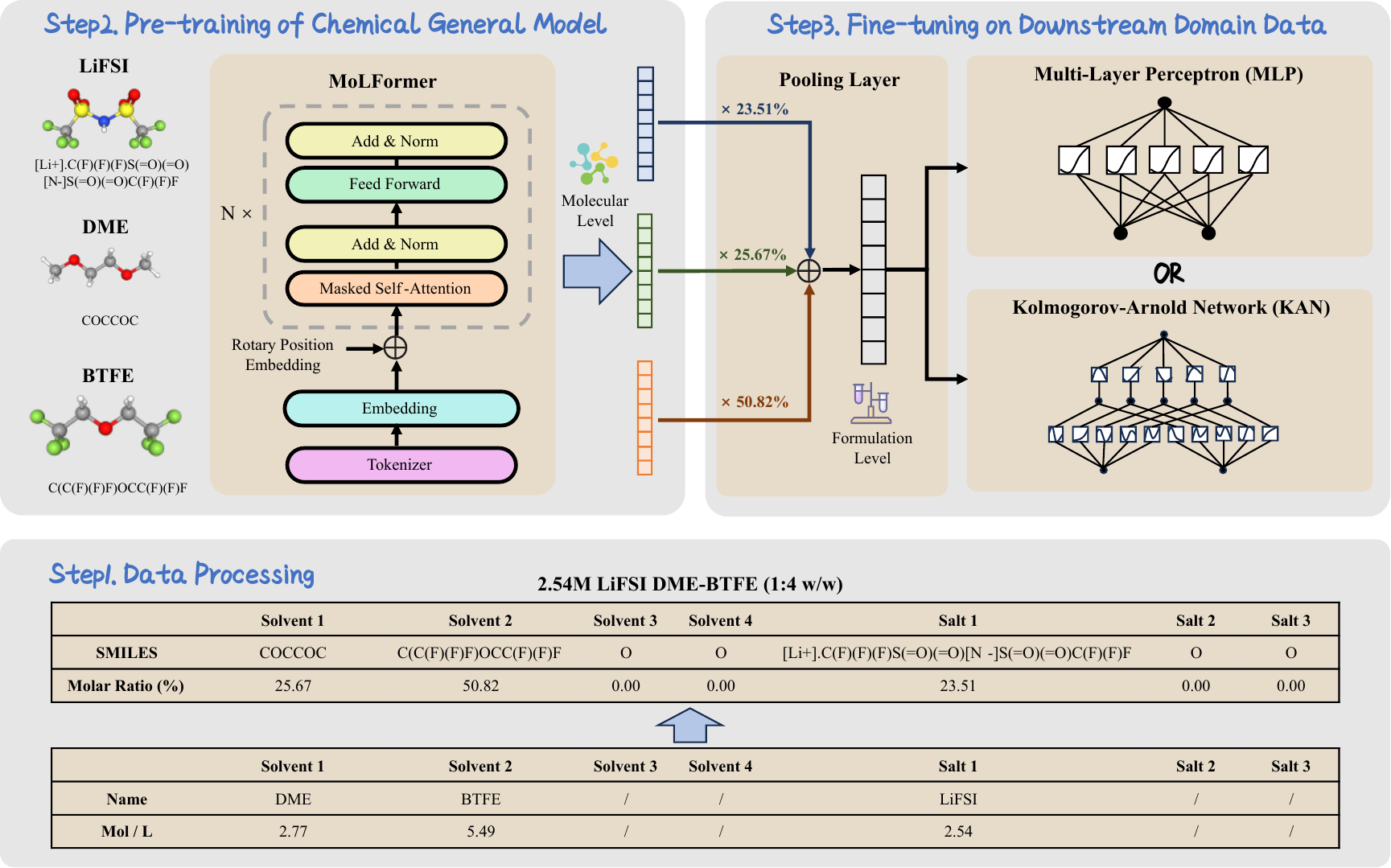}
    \caption{Overview of our proposed \OURMODEL{} framework.}
    \label{fig:method}
\end{figure*}

\subsection{Kolmogorov–Arnold Networks}
Inspired by the Kolmogorov-Arnold representation theorem~\cite{kan-theorem1}, researchers propose Kolmogorov-Arnold Networks (KANs)~\cite{kan} as promising alternatives to MLPs.
Unlike MLPs, KANs have learnable activation functions on weights rather than fixed ones on neurons.
Recent studies have shown that in some specific domains, KANs have advantages in accuracy~\cite{wav-kan,kagnn,gkan}, interpretability~\cite{u-kan,ka2ncd,t-kan}, continuous learning~\cite{ikan}, parameter efficiency~\cite{deepokan,kcn}, etc.
In the field of science, PIKAN~\cite{kan,pikan} combines physics-informed neural networks~\cite{pinn} and KANs to solve a 2D Poison equation, and DeepOKAN~\cite{deepokan} uses a KAN operator network based on Gaussian radial basis functions to solve a 2D orthotropic elasticity problem.
In the field of image processing, U-KAN~\cite{u-kan} redesigns the established U-Net~\cite{u-net} pipeline for medical image segmentation by integrating the dedicated KAN layers on the tokenized intermediate representation, and KCN~\cite{kcn} introduces KANs in various pre-trained convolutional neural network models for remote sensing scene classification tasks.
In the field of human-computer interaction, iKAN~\cite{ikan} is an incremental learning framework for wearable sensor human activity recognition that tackles two challenges simultaneously: catastrophic forgetting and non-uniform inputs.
Inspired by the above works, we apply KANs to the electrolyte field for the first time and obtain better experimental results compared with MLPs.

\section{Methodology}

\subsection{Preliminary}
Our task is to predict the battery's Coulombic Efficiency through the input of electrolyte compositions. 
Suppose an electrolyte has $n$ SMILES molecules $\mathcal{S} = \{s_1, s_2, ..., s_n\}$ and corresponding molar ratios $\mathcal{P} = \{p_1, p_2, ..., p_n\}$. 
Utilizing a pre-trained chemical language model, we can obtain a SMILES embedding set $\mathcal{X} = \{\mathbf{x_1}, \mathbf{x_2}, ..., \mathbf{x_n}\}$, where each $\mathbf{x_i}$ represents the embedding of a single SMILES molecule. 

% \textbf{Input $\mathrm{\textbf{x}}$} is constructed by concatenating each embedding $\mathrm{x_i}$ with its corresponding molar percentage $p_i$
% \begin{equation}
% \mathrm{\textbf{x}} = \sum_{i=1}^{n} \mathrm{x_i} \oplus p_i
% \label{equation:x}
% \end{equation}

% \textbf{Output LCE} is obtained by passing the input $\mathrm{\textbf{x}}$ through downstream networks, including MLP and KAN. CE is computed using LCE.
% \begin{equation}
% \mathrm{CE} = 1 - 10^{(-\mathrm{LCE})}
% \label{equation:LCE}
% \end{equation}

We try two different network structures for the downstream task: 
(1) MLP aims to replicate complex functional mappings via a series of nonlinear transformations across multiple layers. 
The blocks can be formally represented as: 
\begin{equation}
    \mathrm{MLP}(\mathrm{\textbf{x}}) = \left( \textbf{W}_{L-1} \circ \sigma \circ \textbf{W}_{L-2} \circ \ldots \circ \sigma \circ \textbf{W}_0 \right) \mathrm{\textbf{x}}
    \label{equation:MLP}
\end{equation}
where $\textbf{W}_i$ denotes the weight matrix, $\sigma$ denotes the activation function, and $\textbf{x}$ denotes the model input.
(2) KAN is inspired by the Kolmogorov-Arnold representation theorem, and an L-layer KAN operates analogously to an MLP, comprising a sequence of nested KAN layers. 
The blocks can be formally represented as: 
\begin{equation}
    \mathrm{KAN}(\mathrm{\textbf{x}}) = \left( \mathbf{\Phi}_{L-1} \circ \mathbf{\Phi}_{L-2} \circ \ldots \circ \mathbf{\Phi}_{1} \circ \mathbf{\Phi}_{0} \right) \mathrm{\textbf{x}}
    \label{equation:KAN}
\end{equation}
where \(\mathbf{\Phi}_i\) denotes the function matrix corresponding to the $i^{\text{th}}$ KAN layer. 
For each KAN layer with \(n_\text{in}\)-dimensional input and \(n_{\text{out}}\)-dimensional output, \(\mathbf{\Phi}\) can be defined as a matrix of 1D functions:
\begin{equation}
    \mathbf{\Phi} = \{\phi_{q,p}\}
    \quad p = 1,2,\ldots,n_\text{in},  q = 1,2,\ldots,n_{\text{out}}.
\end{equation}

% In summary, KANs are different from regular MLPs because they use learnable activation functions on the edges and parameterized activation functions as weights, which eliminates the need for linear weight matrices. This clever design helps KANs achieve similar or even better performance with smaller model sizes. In addition, their archi tecture makes the models easier to understand while still performing well, which makes them great for various applications.

\subsection{Overview of Method}
Our method \OURMODEL{}, illustrated in the Figure \ref{fig:method}, is built upon an optimized MoLFormer architecture, which is capable of capturing sufficiently precise chemical and structural information solely from SMILES representations. 

First, we employ MoLFormer, a pre-trained Transformer encoder, to generate embeddings for chemical molecules. The dataset comprises a subset from 10\% of both the Zinc and PubChem datasets. We feed each SMILES, $s_i$, individually into MoLFormer to obatain the token embeddings of all SMILES molecules $\mathrm{x_i}$, a vector that represents the chemical features of each component. After that, we scale $\mathrm{x_i}$ by their corresponding molar ratios $p_i$ and subsequently aggregate these weighted embeddings through summation to yield a feature vector that characterizes the entire electrolyte embedding $\mathrm{\textbf{x}}$. This is similar to concatenating all embeddings along the token length dimension, followed by a weighted average pooling layer, with the weights derived from the ratios of the electrolyte components. Finally, we input the electrolyte embedding $\mathrm{\textbf{x}}$ into MLP or KAN to predict the CE of electrolyte compositions.

\subsection{Data Processing}
% The data presented in \ref{cuiyi} are restricted to solvent volume ratios and molar amounts of salts, underscoring the influence of the elemental composition on the electrolyte's coulombic efficiency. It further simplifies the discourse by equating weight ratios with volume ratios, an approximation that might induce a degree of discrepancy in the analytical outcomes. Given this, adopting molar ratios as a means of expressing electrolyte formulations is regarded as a more meticulous approach to replicating experimental conditions.

% Consequently, we have translated the electrolyte formulas into SMILES accompanied by formulation molar ratios, with the intention of bolstering our model's proficiency in forecasting log-coulombic efficiency. This adaptation is instrumental in delving deeper into the intrinsic connections between formulation and battery performance characteristics.

The dataset~\cite{kim} confines attention to solvent volume and salt molarity, emphasizing elemental composition's role in CE. By approximating weight and volume ratios, potential discrepancies are introduced. Thus, adopting molar ratios for formulation enhances accuracy in simulating experimental scenarios.

We converted electrolyte compositions into SMILESs with accompanying molar ratios, better representing the relationship between electrolyte solvents and salts, thereby enhancing the model's capability to predict Logarithmic Coulombic Efficiency (LCE).

% \subsection{Layer}

% In this section, we present the *** component of our proposed approach, wherein each SMILES molecular formula, after being encoded by MoLFormer, is integrated with its respective molar ratios.

% To enhance computational efficiency and optimize resource utilization, the sequence length is meticulously set within a range of 1 to 202 tokens.

% In our SMILES encoding process, we employ a strategic trick: given that MoLFormer's input is constrained to three dimensions and the first dimension solely represents a count with no influence on the encoding process, we concatenate the tokenized SMILES molecular sequences before inputting them. They are later separated post-encoding. Consequently, following MoLFormer encoding for each SMILES formula, the obtained learning vector for each molecule has dimensions of ($B* L * H$), where $N$ signifies the number of SMILES molecule formulas, $B$ represents the batch size, $L$ indicates the number of tokens in each molecule, and every token is embedded into a 768-dimensional feature space. This approach effectively leverages the model's architecture while preserving the integrity of individual molecular representations.

\subsection{Pre-training of Chemical General Model}
Our approach is based on MoLFormer, an advanced transformer-based model commonly utilized for chemical language representations. MoLFormer is a large-scale masked language model that processes inputs using a series of blocks that alternate between self-attention and feed-forward connections.

MoLFormer with linear attention integrates rotary position encoding, demonstrated to outperform absolute position encoding, enhancing the representation of positional relationships in SMILES sequences without significant computational overhead. The attention mechanism is defined as follows:
$$Attention_m(Q,K,V)=\frac{\sum_{n=1}^N\langle \phi(R_mq_m),\phi(R_nq_n)\rangle v_n}{\sum_{n=1}^N\langle \phi(R_mq_m),\phi(R_nq_n)\rangle}$$
where $\phi$ denotes the embedding function, $R$ the rotation matrix, and $Q,K,V$ represent query, key, and value matrices respectively.

Using the capabilities of MoLFormer and enhancing it with the addition of relative positional embeddings, our methodology presents an advanced and efficient solution to anticipate the LCE of battery electrolytes, offering valuable insight into a variety of chemical applications.

\subsection{Fine-tuning on Downstream Domain Data}

We detail the SMILES's tokenization and integration of MoLFormer-encoded SMILES with their molar ratios.

Considering that 99.4 \% of the SMILES sequences comprise no more than 202 tokens, to enhance operational efficiency, we impose a limitation on sequence lengths, setting them within the range of 1 to 202 tokens. 

\textbf{Concatenate Input Disentangle Output (CIDO)} MultiModal-MoLFormer is splicing SMILES through \textless sep\textgreater as input to MoLFormer, which is against the format of MoLFormer pre-training, so we innovate by preprocessing: concatenating tokenised SMILES sequences before input and later disentangling post-encoding. After encoding, the representation of each molecule is a shape tensor $B \times L \times H $, where $B$ is the batch size, $L$ the token count per molecule, and $H$ denotes the embedding dimensionality. This approach capitalises on MoLFormer's design while maintaining the integrity of distinct molecules. design while maintaining the integrity of distinct molecular profiles.

We use two methods: \OURMODELMLP{} and \OURMODELKAN{} as our downstream network structure to predict the CE of the electrolyte formulation. This configuration processes the electrolyte embedding, $\mathrm{\textbf{x}}$, to make the prediction, thereby leveraging the model's capacity to capture intricate relationships within the molecular structures for enhanced efficiency forecasts. The number of layers in the two methods is not fixed. Here we represent the representation obtained by $\mathrm{\textbf{x}}$ after the i-th layer
\begin{equation}
\mathrm{\textbf{x}_{i+1}} = \mathrm{MLP}_{i}(\mathrm{\textbf{x}_{i}}).
\end{equation}
\begin{equation}
\mathrm{\textbf{x}_{i+1}} = \mathrm{KAN}_{i}(W_{i} \times \mathrm{\textbf{x}_{i}} \mid \Phi_{i}).
\end{equation}

\section{Experiments}

\subsection{Experimental Setup}

\subsubsection{Datasets}
\begin{figure}[htbp]
    \centering
    \includegraphics[width=0.9\linewidth]{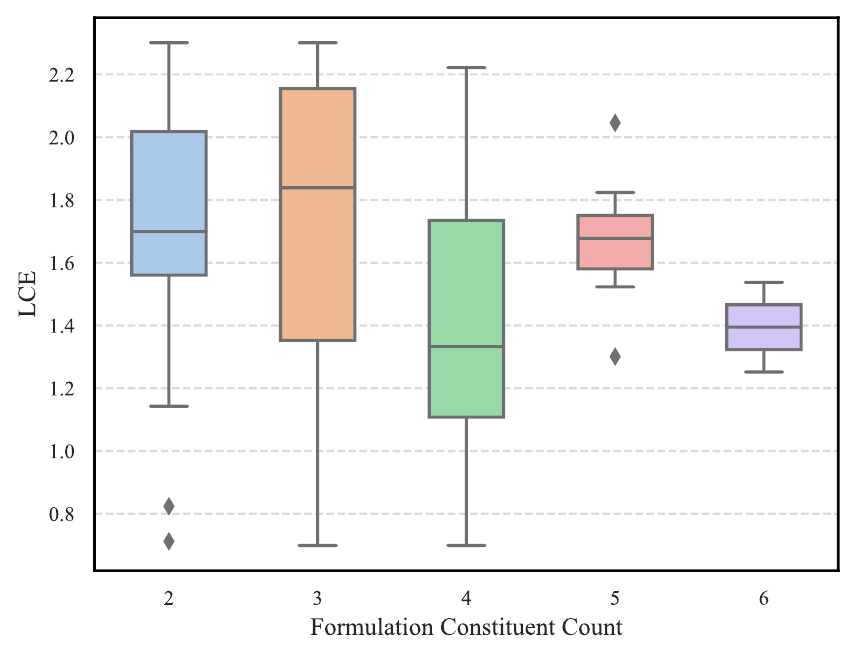}
    \caption{Box plots depict the distribution of LCE outputs for training data based on the formulation constituent count. The outer whiskers represent the minimum and maximum values, the central line represents the median, and the colored box represents the 25th to 75th percentile of the data. Data points outside the outer whiskers are outliers observed in the data.}
    \label{fig:dataset}
\end{figure}

To validate the efficacy of our method, we apply it to the Li/Cu half-cell-based dataset~\cite{kim}, collected from some chemical literature.
Electrolyte formulations in this dataset consist of 2 to 7 components each, described by SMILES.
To numerically amplify the change in output with respect to electrolyte variation, CE is converted to a logarithmic format (LCE, defined as $\mathsf{-log(1 - CE)}$).
We remove an entry with a repeated ratio but different measurement methods, and correct errors in some ratios and molecular information.
Thus, we obtain 149 electrolyte formulations for training and 13 entries as a test set \TESTOUT{}.
Additionally, we split the 149 data into training and test sets in a ratio of 7:3 to obtain \TESTIN{}. 
Box plots illustrated in Figure~\ref{fig:dataset} provide a clear visualization of the distribution of LCE outputs for the training data based on the count of formulation constituents. 

\subsubsection{Baselines}

We compare our method with some strong baselines, which are divided into two groups: machine learning ($\dagger$) and deep learning ($\ddagger$) approaches.

\begin{itemize}
    % \item \textbf{Linear regression} ($\dagger$)~\cite{kim} fits a linear equation to observed data by minimizing the sum of squared residuals between predicted and actual values.
    % \item \textbf{Random forest} ($\dagger$)~\cite{kim} constructs multiple decision trees and combines their outputs to improve prediction accuracy.
    % \item \textbf{Boosting} ($\dagger$)~\cite{kim} combines multiple weak learners to create a strong learner by focusing on the errors of previous models. 
    % \item \textbf{Bagging} ($\dagger$)~\cite{kim} improves model accuracy by combining predictions from multiple models trained on different data subsets.

    \item \textbf{Kim et al.} ($\dagger$)~\cite{kim} use the elemental composition of electrolytes as input features for the machine learning models (e.g., linear regression, random forest, boosting, and bagging) to predict CE.
    \item \textbf{F-GCN} ($\ddagger$)~\cite{f-gcn} assembles multiple Graph Convolution Networks (GCNs) in parallel to intuitively represent formulation constituents, including no-TL F-GCN and TL F-GCN.
    \item \textbf{MoLFormer} ($\ddagger$)~\cite{molformer} trains an efficient Transformer encoder on SMILES sequences of 1.1 billion unlabeled molecules, using a linear attention mechanism and rotational position embedding.
    \item \textbf{MultiModal-MoLFormer} ($\ddagger$)~\cite{multimodal-moLformer} utilizes extensive chemical information learned in pre-training to predict the performance of electrolytes, allowing multiple molecules in SMILES format as input. 
    \item \textbf{BART-SA} ($\ddagger$)~\cite{bart-sa} obtains molecular representations of individual electrolyte components from the pre-trained BART model, then scales and sums to form a unified feature vector.
    \item \textbf{Uni-ELF} ($\ddagger$)~\cite{uni-elf} is a multi-level representation learning framework that enhances electrolyte design through two-stage pre-training.
\end{itemize}

% In this study, we adopt the methodology from the paper "Capturing Formulation Design of Battery Electrolytes with Chemical Large Language Model" as our baseline. This approach utilizes the MoLFormer, a transformer-based chemical language model, to predict the performance of battery electrolyte formulations. The MoLFormer model leverages extensive pre-training on large, unlabeled chemical datasets followed by fine-tuning on specific tasks, effectively capturing the complex relationships within chemical language representations.

% The baseline model inputs up to six SMILES notations along with their respective molar ratios, representing the composition of the electrolyte formulations. The performance of the baseline model is measured using the root mean squared error (RMSE) metric. According to the study, the MoLFormer model achieved an RMSE of 0.195 in predicting the logarithmic Coulombic efficiency (LCE) of electrolyte formulations.

% We use this model and its performance metric as our baseline to compare and evaluate the efficacy of our proposed methodology. By establishing this benchmark, we aim to demonstrate the improvements and advantages of our approach in predicting the performance of battery electrolytes.

\subsubsection{Metrics}
% In this study, we evaluate the performance of our model using Root Mean Square Error (RMSE). RMSE is a widely used metric for regression tasks that measures the average magnitude of the errors between the predicted and observed values. It is calculated as the square root of the average of squared differences between predicted and actual values. Mathematically, RMSE is defined as:
We use Root Mean Square Error (RMSE) as the metric that can be calculated by the following formula:
\begin{equation}
    \text{RMSE} = \sqrt{\frac{1}{N} \sum_{i=1}^{N} (y_i - \hat{y}_i)^2}
\end{equation}
where $N$ is the number of samples, $y_i$ and $\hat{y}_i$ are the experimentally observed and model-predicted LCEs, respectively.
% RMSE is particularly useful as it provides a direct interpretation of the prediction error in the same units as the target variable. A lower RMSE value indicates better predictive accuracy and model performance.By using RMSE, we aim to quantitatively assess the accuracy of our model in predicting the Coulombic efficiency based on the SMILES representation and composition ratio of the electrolyte formulations. This metric allows us to compare different models and configurations to ensure that our proposed approach achieves superior performance.

\subsubsection{Hyperparameters and Training Details}
For \OURMODELMLP{}, we set the dimensions of MoLFormer embedding to 768, headers of attention layers to 12, and the number of encoder layers to 12. 
The downstream MLP consists of 2 linear layers with dropout 0.2 and ReLU~\cite{relu} activation functions.
We optimize the parameters with AdamW~\cite{adamw} with batch size 4, learning rate \(5 \times 10^{-5}\), and epochs 100.
For \OURMODELKAN{}, we set 2 KANs and a linear layer in the downstream network, while other hyperparameters are consistent with \OURMODELMLP{}.

% In our experiments, we fine-tuned the model with the following hyperparameters. The batch size was set to 4, which balances memory constraints and gradient estimation accuracy. We applied a dropout rate of 0.2 to prevent overfitting and improve the generalization capability of the model. The training process was conducted over 100 epochs to ensure convergence and adequate learning of the model parameters. During training, the model's memory usage is approximately 2000MB.

% Our neural network architecture consists of three linear layers, designed to capture both local and global patterns in the data. The optimization was performed using the Adam optimizer, with an initial learning rate of \(5 \times 10^{-5}\). This choice of optimizer and learning rate was made to achieve efficient and stable convergence. Additionally, we employed techniques such as [mention any other techniques, if applicable, e.g., learning rate decay, early stopping, etc.] to further enhance the training process.

% Our KAN network architecture includes two KAN layers followed by a linear layer, tailored to capture intricate local and global patterns within the data. We employed the Adam optimizer with an initial learning rate set at \(3 \times 10^{-5}\), selected to ensure efficient and stable convergence during optimization.

\subsection{Main Experimental Results}
We conduct comparative experiments to evaluate the effectiveness of each method in the predicting LCE task. 
Table~\ref{tab:main} shows the RMSE results of all models on \TESTIN{} and \TESTOUT{} test sets.

% Table generated by Excel2LaTeX from sheet 'main'
\begin{table}[htbp]
  \centering
  \setlength{\tabcolsep}{15pt}
  \caption{RMSE results of LCE prediction for different methods, where bold indicates the best performance and underline denotes the second-best performance. * stands for the ratio of training set to test set is 8:2, so it is not included in the comparison. }
  \resizebox{0.9\linewidth}{!}{
    \begin{tabular}{ccc}
    \toprule
    \multirow{2}[4]{*}{\textbf{Methods}} & \multicolumn{2}{c}{\textbf{LCE}} \\
\cmidrule{2-3}          & \TESTIN{} & \TESTOUT{} \\
    \midrule
    Linear Regression & 0.585 & 0.170 \\
    Random Forest & 0.577 &  - \\
    Boosting & 0.587 &  - \\
    Bagging & 0.583 &  - \\
    \midrule
    no-TL F-GCN &  -    & 0.779 \\
    TL F-GCN &  -    & 0.390 \\
    MoLFormer &  -    & 0.213 \\
    MultiModal-MoLFormer &  -    & 0.195 \\
    BART-SA* & 0.148 &  - \\
    Uni-ELF & \textbf{0.184} &  - \\
    \midrule
    \textbf{COEFF-MLP} & 0.221     & \underline{0.132} \\
    \textbf{COEFF-KAN} & \underline{0.186}     & \textbf{0.110} \\
    \bottomrule
    \end{tabular}%
    }
  \label{tab:main}%
\end{table}%

First, we observe that our method outperforms machine learning models on all datasets. 
Especially on \TESTIN{}, the RMSE of \OURMODELKAN{} is 0.399, 0.391, 0.401, and 0.397 lower than linear regression, random forest, boosting, and bagging respectively.
The performance of these approaches is limited by whether the feature selection is appropriate and complete.
For example, \citet{kim} use the elemental composition of liquid electrolytes as input features.
They successfully reveal that a reduction in the solvent oxygen content is critical for superior CE but overlook structural information of chemical molecules (e.g., geometric arrangements of elements in isomers). 
This subtle distinction has been shown to affect electrolyte properties and battery performance~\cite{structure1,structure2}.

\begin{figure}[htbp]
    \centering
    \includegraphics[width=0.9\linewidth]{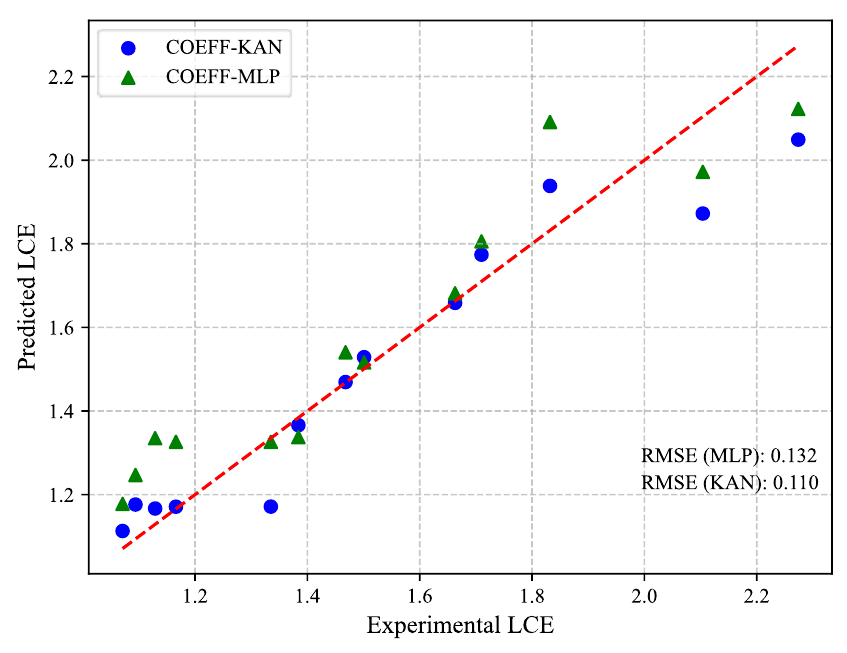}
    \caption{Parity plots show predicted LCE values as scatterplots with respect to the actual values.}
    \label{fig:main}
\end{figure}

Second, it is noticed that our method achieves state-of-the-art performance among all deep learning models on \TESTOUT{}.
In particular, the RMSE of \OURMODELKAN{} is 0.280, 0.103, and 0.085 lower than F-GCN, MoLFormer, and MultiModal-MoLFormer.
Although our method, like most others, follows a two-stage paradigm, it benefits from the accurate representation of chemical molecules by MoLFormer, weighted average pooling at the formulation level, and performance advantages of KANs in the scientific domain.
Additionally, unlike MultiModal-MoLFormer, which uses data augmentation to expand the training set to 27,266, our model only requires fine-tuning with just over a hundred samples to achieve more accurate predictions.

Third, experimental results demonstrate that introducing KANs into downstream networks can better predict the LCE of electrolytes.
Specifically, the RMSE of \OURMODELKAN{} is 0.030 and 0.022 lower than that of \OURMODELMLP{} on \TESTIN{} and \TESTOUT{}, respectively.
We speculate that there are two main reasons: 
(1) Better interpretability: Compared to MLPs, design of KANs network structure makes the model's behavior easier to understand. 
By observing the external connection patterns, we can have an insight into interactions between input features.
The internal activation functions help visualize how a single variable affects the output, thereby increasing the transparency of models.
(2) More efficient learning with limited data size: Theoretically, KANs can achieve better fitting effects with fewer samples, because they learn more complex and generalizable function mappings through the synergy of trainable network structures and activation functions.

Finally, our method is stable and effective for predicting LCE against unseen data.
As shown in Figure~\ref{fig:main}, most of the predicted data points for \OURMODELMLP{} and \OURMODELKAN{} are close to the perfect prediction line and have similar distributions.
The generalization can make \OURMODEL{} adapt to practical application scenarios and help chemical researchers reduce the workload of experimental verification.

\subsection{Ablation Experiments}

In this section we will perform ablation experiments and analyses in terms of CIDO and KAN respectively

\textbf{CIDO.} 
% 表1报告了是否使用该方法的消融情况。在LCE prediction的任务上，使用CIDO方法$T_{out}$提升了xxx的效果。结果表明，使用CIDO对模型预测是有效的，与此同时，在相同的机器和配置下，训练时间方面也节省了接近三倍开销，大大减少了训练的时间，如表2所示。
Table~\ref{tab:CIDO} reports the ablation with or without the method. In the LCE prediction task, MLP and KAN use the CIDO method to improve performance on $T_{out}$ by 0.261 and 0.278, respectively. The results show that using CIDO can be effective for model prediction while saving some training time overhead with the same machine and configuration.
\begin{table}[htbp]
\centering
\small
{%
\begin{tabular}{ccc}
\hline
\textbf{Methods} & \TESTOUT{} & \textbf{Time}(s)\\
\hline
\OURMODELMLP{} w/o CIDO & 0.393 & 715 \\
\OURMODELMLP{} w/ CIDO & 0.132 & 705 \\
\OURMODELKAN{} w/o CIDO & 0.388 & 741  \\
\OURMODELKAN{} w/ CIDO & 0.110 & 725 \\
\hline
\end{tabular}
}
      \caption{RMSE results of LCE prediction at w/ and w/o CIDO method.}
\label{tab:CIDO}
\end{table}

\begin{figure}[htbp]
    \centering
    \includegraphics[width=\linewidth]{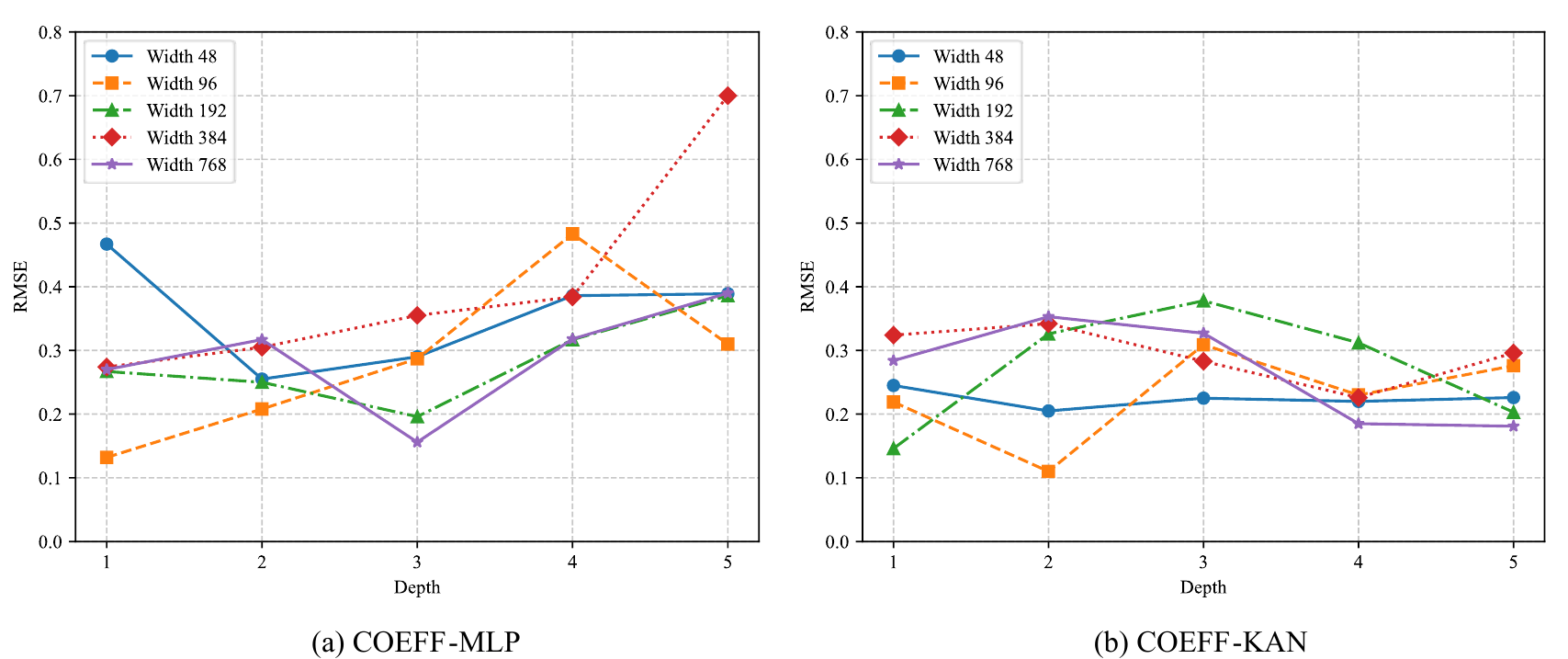}
    \caption{RMSE results of LCE prediction at different network depths and widths.}
    \label{fig:ablation}
\end{figure}

\textbf{KAN.} 
% 我们用不同的深度和宽度的MLP作为基线与KAN来比较， MLP与KAN均使用了CIDO进行了100epochs的训练，将验证集的最低loss的epoch在测试集的RMSE绘制为Figure  \ref{Figure:5}。 从图上可知，在T_out任务上进行测试，我们发现KAN的最优值比MLP好17.42\%, 从而证明下游网络采用KAN效性。与此同时，我们发现不同宽度的KAN的结果随着网络深度RMSE趋于稳定，但是MLP在2-3层网络架构的时候性能较优，随着网络的深度加深，性能逐渐不稳定。
We used MLP with different depths and widths as a baseline to compare with KAN. Both MLP and KAN were trained with 100 epochs using CIDO, and the RMSE of the lowest-loss epoch of the validation set on the test set is plotted as Figure \ref{fig:ablation}. As can be seen from the figure, testing on the $T_{out}$, we find that the optimal value of KAN is 17.42\% better than that of MLP, thus proving the effectiveness of the downstream network using KAN. At the same time, we find that the results of KAN with different widths tend to stabilize with the network depth RMSE, but MLP performs better at 2-3 layers of the network architecture, and the performance is gradually unstable as the depth of the network deepens.

% \subsection{Case Study}

\section{Conclusion and Future Work}

In this paper, we propose a novel method \OURMODEL{} for accurately predicting the LCE of electrolytes.
There are two main contributions: 

(1) We adopt a two-stage paradigm: pre-training a chemical general model and fine-tuning it on domain data. 

(2) We innovatively replace the downstream MLPs with KANs.

In the experiments, we show that our method significantly and consistently outperforms state-of-the-art models on a real-world test set \TESTOUT{}.
% 虽然KAN 可能会过拟合，特别是在数据有限的情况下。 它们形成复杂模型的能力可能会将噪声误认为是重要模式，导致泛化能力差。但是我们在保证域内能力的同时，域外能力也有很好的性能。
% 增强数据集

In the future, we will explore the following directions:

(1) We will apply our model to more electrolyte properties (e.g., charge/discharge current and battery cycle life), more chemical fields, and even material science subfields, as the two-stage paradigm is universal.

(2) We will fully utilize the interpretability of KANs to find the quantitative relationship between electrolyte formulations and CE through function fitting and visualization.
% (1) 需要修改，我们将应用我们的模型来预测更多的电解液性质，更多的化学领域，甚至物质科学子领域性质预测，
% (2) 我们将充分利用KAN的可解释性，通过函数拟合和可视化找到电解质配方与CE之间的定量关系。

\bibliography{aaai25}

\end{document}